\pdfoutput=1
\documentclass[11pt,a4paper]{article}
\usepackage[hyperref]{naaclhlt2018}
\usepackage{times}
\usepackage{graphicx}
\usepackage{url}

\usepackage{amsmath}
\usepackage{amssymb}
\usepackage{mathtools}

\usepackage{ifthen}
\usepackage{tikz}
\usetikzlibrary{arrows}
\usetikzlibrary{shapes.geometric}
\usepackage{pgfplots}
\pgfplotsset{compat=1.3}

\DeclareMathOperator*{\argmax}{argmax}

\aclfinalcopy 

\title{What's Going On in Neural Constituency Parsers? An Analysis}

\author{David Gaddy, Mitchell Stern, and Dan Klein \\
  Computer Science Division \\
  University of California, Berkeley \\
  {\tt \{dgaddy,mitchell,klein\}@berkeley.edu} \\}

\date{}

\begin{document}

\maketitle

\begin{abstract}

A number of differences have emerged between modern and classic approaches to constituency parsing in recent years, with structural components like grammars and feature-rich lexicons becoming less central while recurrent neural network representations rise in popularity. The goal of this work is to analyze the extent to which information provided directly by the model structure in classical systems is still being captured by neural methods. To this end, we propose a high-performance neural model (92.08 F1 on PTB) that is representative of recent work and perform a series of investigative experiments. We find that our model implicitly learns to encode much of the same information that was explicitly provided by grammars and lexicons in the past, indicating that this scaffolding can largely be subsumed by powerful general-purpose neural machinery.

\end{abstract}

\section{Introduction}

In the past several years, many aspects of constituency parsing and natural language processing in general have changed.
Grammars, which were once the central component of many parsers, have played a continually decreasing role.
Rich lexicons and handcrafted lexical features have become less common as well.
On the other hand, recurrent neural networks have gained traction as a powerful and general purpose tool for representation.
So far, not much has been shown about
how neural networks are able to compensate for the removal of the structures used in past models.
To gain insight, we introduce a parser that is representative of recent trends and analyze its learned representations to determine what information it captures and what is important for its strong performance.

Our parser is a natural extension of recent work in constituency parsing. We combine a common span representation based on recurrent neural networks with a novel, simplified scoring model. In addition, we replace the externally predicted part-of-speech tags used in some recent systems with character-level word representations. Our parser achieves a test F1 score of 92.08 on section 23 of the Penn Treebank, exceeding the performance of many other state-of-the-art models evaluated under comparable conditions. Section~\ref{sec:model} describes our model in detail.

The remainder of the paper is focused on analysis.
In Section~\ref{sec:grammar}, we look at the decline of grammars and output correlations.  Past work in constituency parsing used context-free grammars with production rules governing adjacent labels (or more generally production-factored scores) to propagate information and capture correlations between output decisions \cite{collins1997three,charniak2005coursetofine,petrov2007improved,hall2014less}.
Many recent parsers no longer have explicit grammar production rules, but still use information about other predictions, allowing them to capture output correlations \cite{dyer2016rnng,choe2016parsing}.
Beyond this, there are some parsers that use no context for bracket scoring and only include mild output correlations in the form of tree constraints \cite{cross2016span,stern2017minimal}.
In our experiments, we find that we can accurately predict parents from the representation given to a child.  Since a simple classifier can predict the information provided by parent-child relations, this explains why the information no longer needs to be specified explicitly.
We also show that we can completely remove output correlations from our model with a variant of our parser that makes independent span label decisions without any tree constraints while maintaining high F1 scores and mostly producing trees.

In Section~\ref{sec:lex}, we look at lexical representations.
In the past, parsers used a variety of custom lexical representations, such as word shape features, prefixes, suffixes, and special tokens for categories like numerals \cite{klein2003accurate,petrov2007improved,finkel2008efficient}.
Character-level models have shown promise in parsing and other NLP tasks as a way to remove the complexity of these lexical features \cite{ballesteros2015improved,ling2015character,kim2016character,coavoux2017multilingual,liu2017shiftreduce}.
We compare the performance of character-level representations and externally predicted part-of-speech tags and show that these two sources of information seem to fill a similar role.  We also perform experiments showing that the representations learned with character-level models contain information that was hand-specified in some other models.

Finally, in Section~\ref{sec:distance} we look at the surface context captured by recurrent neural networks. Many recent parsers use LSTMs, a popular type of recurrent neural network, to combine and summarize context for making decisions \cite{choe2016parsing,cross2016incremental,dyer2016rnng,stern2017minimal}.  Before LSTMs became common in parsing, systems that included surface features used a fixed-size window around the fenceposts at each end of a span \cite{charniak2005coursetofine,finkel2008efficient,hall2014less,durrett2015neural}, and the inference procedure handled most of the propagation of information from the rest of the sentence.  We perform experiments showing that LSTMs capture far-away surface context and that this information is important for our parser's performance.
We also provide evidence that word order of the far-away context is important and that the amount of context alone does not account for all of the gains seen with LSTMs.

Overall, we find that the same sources of information that were effective for grammar-driven parsers are also captured by parsers based on recurrent neural networks.

\section{Parsing Model}
\label{sec:model}

In this section, we propose a span-based parsing model that combines components from several recent neural architectures for constituency parsing and other natural language tasks. While this system is primarily introduced for the purpose of our analysis, it also performs well as a parser in its own right, exhibiting some gains over comparable work. Our model is in many respects similar to the chart parser of \citet{stern2017minimal}, but features a number of simplifications and improvements.

\subsection{Overview}

Abstractly, our model consists of a single scoring function $s(i,j,\ell)$ that assigns a real-valued score to every label $\ell$ for each span $(i,j)$ in an input sentence. We take the set of available labels to be the collection of all nonterminals and unary chains observed in the training data, treating the latter as atomic units.
The score of a tree $T$ is defined as a sum over internal nodes of labeled span scores: $$s(T) = \sum_{(i,j,\ell) \in T} s(i,j,\ell).$$ We note that, in contrast with many other chart parsers, our model can directly score $n$-ary trees without the need for binarization or other tree transformations.
Under this setup, the parsing problem is to find the tree with the highest score: $$\hat{T} = \argmax_T s(T).$$ Our concrete implementation of $s(i,j,\ell)$ can be broken down into three pieces: word representation, span representation, and label scoring. We discuss each of these in turn.

\subsection{Word Representation}

One popular way to represent words is the use of word embeddings. We have a separate embedding for each word type in the training vocabulary and map all unknown words at test time to a single {\tt <UNK>} token.
In addition to word embeddings, character-level representations have also been gaining traction in recent years, with common choices including recurrent, convolutional, or bag-of-$n$-gram representations. These alleviate the unknown word problem by working with smaller, more frequent units, and readily capture morphological information not directly accessible through word embeddings. Character LSTMs in particular have proved useful in constituency parsing \citep{coavoux2017multilingual}, dependency parsing \citep{ballesteros2015improved}, part-of-speech tagging \citep{ling2015finding}, named entity recognition \citep{lample2016neural}, and machine translation \citep{ling2015character}, making them a natural choice for our system. We obtain a character-level representation for a word by running it through a bidirectional character LSTM and concatenating the final forward and backward outputs.

The complete representation of a given word is the concatenation of its word embedding and its character LSTM representation. While past work has also used sparse indicator features \citep{finkel2008efficient} or part-of-speech tags predicted by an external system \citep{cross2016span} for additional word-level information, we find these to be unnecessary in the presence of a robust character-level representation.

\subsection{Span Representation}

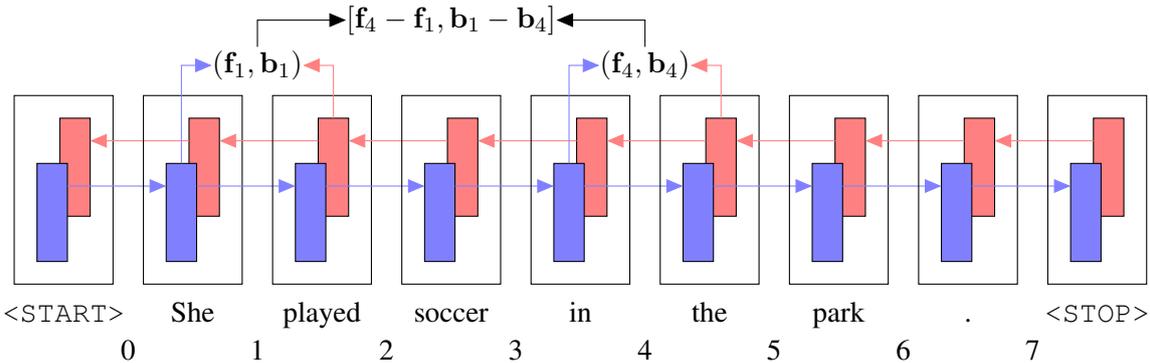
\begin{figure*}
\centering
\begin{tikzpicture}[every node/.style={font=\strut,inner sep=0},>=triangle 45]
\newcommand\sentence{{\tt<START>},She,played,soccer,in,the,park,.,{\tt<STOP>}}
\newcommand\leftIndex{1}
\newcommand\rightIndex{4}
\newcommand\forwardColor{white!50!blue}
\newcommand\backwardColor{white!50!red}
\newcommand\lstmPosition{(0,0)}
\newcommand\lstmIndex{0}
\pgfmathsetmacro{\wordDistance}{1.7}
\pgfmathsetmacro{\indexOffsetV}{-.5}
\pgfmathsetmacro{\forwardOffsetV}{.7}
\pgfmathsetmacro{\lstmWidth}{.4}
\pgfmathsetmacro{\lstmHeight}{1.3}
\pgfmathsetmacro{\lstmSeparationH}{.3}
\pgfmathsetmacro{\lstmSeparationV}{.6}
\pgfmathsetmacro{\forwardArrowOffsetV}{\lstmHeight-\lstmSeparationV/2}
\pgfmathsetmacro{\backwardArrowOffsetV}{\lstmHeight-\lstmSeparationV/2}
\pgfmathsetmacro{\boxMargin}{.3}
\pgfmathsetmacro{\outputOffsetV}{.4}
\pgfmathsetmacro{\spanOffsetV}{.6}
\foreach \word [count=\finalIndex from -1] in \sentence;
\foreach \word [count=\i from -1] in \sentence {
  \path \lstmPosition ++ ({\wordDistance*(\i)},0)
    node (word\i) {\word};
  \ifthenelse{\i>-1}{
    \node at ({\wordDistance*(\i-0.5)},\indexOffsetV) {\i};
  }{}
  \draw[fill=\backwardColor]
    \lstmPosition++(\wordDistance*\i-\lstmWidth/2+\lstmSeparationH/2,\forwardOffsetV+\lstmSeparationV)
    node (backwardLowerLeft) {}
    rectangle ++(\lstmWidth,\lstmHeight)
    node[midway] (backwardCenter) {}
    node (backwardUpperRight) {};
  \draw[fill=\forwardColor]
    \lstmPosition++(\wordDistance*\i-\lstmWidth/2-\lstmSeparationH/2,\forwardOffsetV)
    node (forwardLowerLeft) {}
    rectangle ++(\lstmWidth,\lstmHeight)
    node[midway] (forwardCenter) {}
    node (forwardUpperRight) {};
  \path
    (forwardCenter |- forwardUpperRight)
    node (forwardOut\i) {};
  \path
    (backwardCenter |- backwardUpperRight)
    node (backwardOut\i) {};
  \ifthenelse{\i>-1}{
    \draw[<-,\forwardColor]
      (forwardLowerLeft)
      ++(0,\forwardArrowOffsetV)
      -- ++(-\wordDistance+\lstmWidth,0);
    \draw[->,\backwardColor]
      (backwardLowerLeft)
      ++(0,\backwardArrowOffsetV)
      -- ++(-\wordDistance+\lstmWidth,0);
  }{}
  \path
    (forwardLowerLeft) ++(-\boxMargin,-\boxMargin)
    node (boxLowerLeft) {};
  \path
    (backwardUpperRight) ++(\boxMargin,\boxMargin)
    node (boxUpperRight) {};
  \draw
    (boxLowerLeft) rectangle (boxUpperRight)
    node[midway] (boxCenter) {};
  \ifthenelse{\i=\leftIndex \OR \i=\rightIndex}{
    \path
      (boxCenter |- boxUpperRight) ++(-\wordDistance*0.5,\outputOffsetV)
      node (output\i) {$(\mathbf{f}_\i,\mathbf{b}_\i)$};
    \pgfmathtruncatemacro{\iPrev}{\i-1}
    \draw[->,\forwardColor] (forwardOut\iPrev.center) |- (output\i);
    \draw[->,\backwardColor] (backwardOut\i.center) |- (output\i);
  }{}
}
\path
  (output\leftIndex) -- (output\rightIndex)
  node[midway] (outputMidpoint) {};
\path
  (outputMidpoint) ++(0,\spanOffsetV)
  node (span) {$[\mathbf{f}_\rightIndex-\mathbf{f}_\leftIndex, \mathbf{b}_\leftIndex-\mathbf{b}_\rightIndex]$};
\draw[->] (output\leftIndex) |- (span);
\draw[->] (output\rightIndex) |- (span);
\end{tikzpicture}
\caption{Span representations are computed by running a bidirectional LSTM over the input sentence and taking differences of the output vectors at the two endpoints. Here we illustrate the process for the span $(1,4)$ corresponding to ``played soccer in'' in the example sentence.}
\label{fig:span-representation}
\end{figure*}

To build up to spans, we first run a bidirectional LSTM over the sequence of word representations for an input sentence to obtain context-sensitive forward and backward representations $\mathbf{f}_i$ and $\mathbf{b}_i$ for each fencepost $i$. We then follow past work in dependency parsing \citep{wang2016graph} and constituency parsing \citep{cross2016span,stern2017minimal} in representing the span $(i,j)$ by the concatenation of the corresponding forward and backward span differences: $$\mathbf{r}_{ij} = [\mathbf{f}_j-\mathbf{f}_i, \mathbf{b}_i-\mathbf{b}_j].$$
See Figure~\ref{fig:span-representation} for an illustration.

\subsection{Label Scoring}

Finally, we implement the label scoring function by feeding the span representation through a one-layer feedforward network whose output dimensionality equals the number of possible labels. The score of a specific label $\ell$ is the corresponding component of the output vector: $$s(i,j,\ell) = \left[ \mathbf{W}_2 \, g(\mathbf{W}_1 \mathbf{r}_{ij} + \mathbf{z}_1) + \mathbf{z}_2 \right]_\ell,$$ where $g$ is an elementwise ReLU nonlinearity.

\subsection{Inference}

Even though our model operates on $n$-ary trees, we can still employ a CKY-style algorithm for efficient globally optimal inference by introducing an auxiliary empty label $\varnothing$ with $s(i,j,\varnothing) = 0$ for all $(i,j)$ to handle spans that are not constituents.
Under this scheme, every binarization of a tree with empty labels at intermediate dummy nodes will have the same score, so an arbitrary binarization can be selected at training time with no effect on learning.
We contrast this with the chart parser of \citet{stern2017minimal}, which assigns different scores to different binarizations of the same underlying tree and in theory may exhibit varying performance depending on the method chosen for conversion.

With this change in place, let $s_\mathrm{best}(i,j)$ denote the score of the best subtree spanning $(i,j)$. For spans of length one, we need only consider the choice of label: $$s_\mathrm{best}(i,i+1) = \max_\ell s(i,i+1,\ell).$$ For general spans $(i,j)$, we have the following recursion:
\begin{align*}
s_\mathrm{best}(i,j) &= \max_\ell s(i,j,\ell) \\
&+ \max_k \left[ s_\mathrm{best}(i,k) + s_\mathrm{best}(k,j) \right].
\end{align*}
That is, we can independently select the best label for the current span and the best split point, where the score of a split is the sum of the best scores for the corresponding subtrees.

To parse the full sentence, we compute $s_\mathrm{best}(0,n)$ using a bottom-up chart decoder, then traverse backpointers to recover the tree achieving that score. Nodes assigned the empty label are omitted during the reconstruction process to obtain the full $n$-ary tree. The overall complexity of this approach is $\mathcal{O}(n^3 + L n^2)$, where $n$ is the number of words and $L$ is the total number of labels. We note that because our system does not use a grammar, there is no constant for the number of grammar rules multiplying the $\mathcal{O}(n^3)$ term as in traditional CKY parsing. In practice, the $\mathcal{O}(n^2)$ evaluations of the span scoring function corresponding to the $\mathcal{O}(Ln^2)$ term dominate runtime.


\subsection{Training}

As is common for structured prediction problems \citep{taskar2005learning}, we use margin-based training to learn a model that satisfies the constraints $$s(T^*) \ge s(T) + \Delta(T,T^*)$$ for each training example, where $T^*$ denotes the gold output, $T$ ranges over all valid trees, and $\Delta$ is the Hamming loss on labeled spans. Our training objective is the hinge loss: $$\max \left( 0, \, \max_T \left[ s(T) + \Delta(T,T^*) \right] - s(T^*) \right).$$ This is equal to 0 when all constraints are satisfied, or the magnitude of the largest margin violation otherwise.

Since $\Delta$ decomposes over spans, the inner loss-augmented decode $\max_T \left[ s(T) + \Delta(T,T^*) \right]$ can be performed efficiently using a slight modification of the dynamic program used for inference. In particular, we replace $s(i,j,\ell)$ with $s(i,j,\ell) + 1[\ell \ne \ell_{ij}^*]$, where $\ell_{ij}^*$ is the label of span $(i,j)$ in the gold tree $T^*$.

\subsection{Results}
\label{subsec:model-implementation}

We use the Penn Treebank \citep{marcus93building} for our experiments with the standard splits of sections 2-21 for training, section 22 for development, and section 23 for testing. Details about our model hyperparameters and training prodecure can be found in Appendix~\ref{appendix:hyperparameters}.

Across 10 trials, our model achieves an average development F1 score of 92.22 on section 22 of the Penn Treebank. We use this as our primary point of comparison in all subsequent analysis. The model with the best score on the development set achieves a test F1 score of 92.08 on section 23 of the Penn Treebank, exceeding the performance of other recent state-of-the-art discriminative models which do not use external data or ensembling.\footnote{Code for our parser is available at \url{https://github.com/dgaddy/parser-analysis}.}

\section{Output Correlations}
\label{sec:grammar}

Output correlations are information about compatibility between outputs in a structured prediction model.
Since outputs are all a function of the input, output correlations are not necessary for prediction when a model has access to the entire input.
In practice, however, many models throughout NLP have found them useful \cite{collins1997three,lafferty2001conditional,koo2010efficient}, and \citet{liang2008structure} provides theoretical results suggesting they may be useful for learning efficiently.
In constituency parsing, there are two primary forms of output correlation typically captured by models.
The first is correlations between label decisions, which often are captured by either production scores or the history in an incremental tree-creation procedure.
The second, more subtle correlation comes from the enforcement of tree constraints, since the inclusion of one bracket can affect whether or not another bracket can be present.
We explore these two classes of output correlations in Sections~\ref{sec:parentclass} and \ref{sec:independent} below.

\subsection{Parent Classification}
\label{sec:parentclass}

The base parser introduced in Section~\ref{sec:model} scores labeled brackets independently then uses a dynamic program to select a set of brackets that forms the highest-scoring tree.  This independent labeling is an interesting departure from classical parsing work where correlations between predicted labels played a central role.
It is natural to wonder why modeling label correlations isn't as important as it once was.  Is there something about the neural representation that allows us to function without it?
One possible explanation is that the neural machinery, in particular the LSTM, is handling much of the 
reconciliation between labels that was previously handled by an inference procedure.
In other words, instead of using local information to suggest several brackets and letting the grammar handle interactions between them, the LSTM may be making decisions about brackets already in its latent state, allowing it to use the result of these decisions to inform other bracketings.

One way to explore this hypothesis would be to evaluate whether the parser's learned representations could be used to predict parent labels of nodes in the tree.
If the label of a node's parent can be predicted with high accuracy from the representation of its span, then little of the information about parent-child relations provided explicitly by a grammar has been lost.
For this experiment, we freeze the input and LSTM parameters of our base model and train a new label scoring network to predict the label of a span's parent rather than the label of the span itself.
We only predict parent labels for spans that have a bracket in the gold tree, so that all but the top level spans will have non-empty labels. The new network is trained with a margin loss.

After training on the standard training sections of the treebank, the network was able to correctly predict 92.3\% of parent labels on the development set.
This is fairly accurate, which supports the hypothesis that the representation knows a substantial amount about surrounding context in the output tree.
For comparison, given only a span's label, the best you can do for predicting the parent is 43.3\% with the majority class conditioned on the current label.

\subsection{Independent Span Decisions}
\label{sec:independent}

Like other recent parsers that do not capture correlations between output labels \cite{cross2016span,stern2017minimal}, our base parser still does have some output correlations captured by the enforcement of tree constraints.  In this section, we set out to determine the importance of these output correlations by making a version of the parser where they are removed.
Although parsers are typically designed to form trees, the bracketing F1 measure used to evaluate parsers is still defined on non-tree outputs.
To remove all output correlations from our parser, we can simply remove the tree constraint and independently make decisions about whether to include a bracketed span.
The architecture is identical to the one described in Section~\ref{sec:model}, producing a vector of label scores for each span.
We choose the label with the maximum score as the label for a span.
As before, we fix the score of the empty label at zero, so if all other label scores are negative, the span will be left out of the set of predicted brackets.
We train with independent margin losses for each span.

Ignoring tree well-formedness, the development F1 score of this independent span selection parser is 92.20, effectively matching the performance of the tree-constrained parser. In addition, we find that 94.5\% of predicted bracketings for development set examples form valid trees, even though we did not explicitly encourage this.
This high performance shows that our parser can function well even without modeling any output correlations.

\section{Lexical Representation}
\label{sec:lex}

In this section, we investigate several common choices for lexical representations of words and their role in neural parsing.

\subsection{Alternate Word Representations}

We compare the performance of our base model, which uses word embeddings and a character LSTM, with otherwise identical parsers that use other combinations of lexical representations.
The results of these experiments are summarized in Table \ref{table:lex}.
First, we remove the character-level representations from our model, leaving only the word embeddings. We find that development performance drops from 92.22 F1 to 91.44 F1, showing that word embeddings alone do not capture sufficient information for state-of-the-art performance. Then, we replace the character-level representations with embeddings of part-of-speech tags predicted by the Stanford tagger \citep{toutanova2003tagging}. This model achieves a comparable development F1 score of 92.09, but unlike our base model relies on outputs from an external system. Next, we train a model which includes all three lexical representations: word embeddings, character LSTM representations, and part-of-speech tag embeddings. We find that development performance is nearly identical to the base model at 92.24 F1, suggesting that character representations and predicted part-of-speech tags provide much of the same information.
Finally, we remove word embeddings and rely completely on character-level embeddings.  After retuning the character LSTM size, we find that a slightly larger character LSTM can make up for the loss in word-level embeddings, giving a development F1 of 92.24.

\begin{table}
\centering
\begin{tabular}{|l|c|}
\hline
Word and Character LSTM & 92.22 \\ \hline
Word Only & 91.44 \\ \hline
Word and Tag & 92.09 \\ \hline
Word, Tag, and Character LSTM & 92.24 \\ \hline
Character Only & 92.24 \\ \hline
\end{tabular}
\caption{Development F1 scores on section 22 of the Penn Treebank for different lexical representations.}
\label{table:lex}
\end{table}

\subsection{Predicting Word Features}
\label{sec:wordfeat}

Past work in constituency parsing has demonstrated that indicator features on word shapes, suffixes, and similar attributes provide useful information beyond the identity of a word itself, especially for rare and unknown tokens \citep{finkel2008efficient,hall2014less}. We hypothesize that the character-level LSTM in our model learns similar information without the need for manual supervision. To test this, we take the word representations induced by the character LSTM in our parser as fixed word encodings, and train a small feedforward network to predict binary word features defined in the Berkeley Parser \citep{petrov2007improved}. We randomly split the vocabulary of the Penn Treebank into two subsets, using 80\% of the word types for training and 20\% for testing.

We find that the character LSTM representations allow for previously handcrafted indicator features to be predicted with accuracies of 99.7\% or higher in all cases. The fact that this simple classifier performs so well indicates that the information contained in these features is readily available from our model's character-level encodings. A detailed breakdown of accuracy by feature can be found in Appendix~\ref{appendix:charlstm}.

\section{Context in the Sentence LSTM}
\label{sec:distance}

In this section, we analyze where the information in the sentence-level LSTM hidden vectors comes from.
Since the LSTM representations we use to make parsing decisions come from the fenceposts on each side of a span, we would like to understand whether they only capture information from the immediate vicinity of the fenceposts or if they also contain more distant information.
Although an LSTM is theoretically capable of incorporating an arbitrarily large amount of context, it is unclear how much context it actually captures and whether this context is important for parsing accuracy.

\subsection{Derivative Analysis}

First, we would like to know if the LSTM features capture distant information.
For this experiment, we use derivatives as a measure of sensitivity to changes in an input.
If the derivative of a value with respect to a particular input is high, then that input has a large impact on the final value.
For a particular component of an LSTM output vector, we compute its gradient with respect to each LSTM input vector, calculate the $\ell_2$-norms of the gradients, and bucket the results according to distance from the output position. This process is repeated for every output position of each sentence in the development set, and the results are averaged within each bucket. Due to the scale of the required computation, we only use a subset of the output vector components to compute the average, sampling one at random per output vector.

\begin{figure}
\centering
\begin{tikzpicture}
\begin{axis}[
  width=\linewidth,
  height=.6\linewidth,
  ybar,
  bar width=2pt,
  xlabel=Distance,
  ylabel=Average Derivative,
  xmin=0,
  xmax=41,
  skip coords between index={40}{999},
  ymin=0,
  yticklabel style={
    /pgf/number format/fixed,
    /pgf/number format/fixed zerofill,
    /pgf/number format/precision=2},
]
\addplot[fill=green] table[x expr=\coordindex+1,y index=0] {
0.1205393148
0.0845026698
0.0656451374
0.0533884146
0.0448505053
0.0377917352
0.0324590923
0.0282639455
0.0250051015
0.0225220263
0.0203399791
0.0186301515
0.0174101097
0.0160527995
0.0151214849
0.0144965892
0.0137872321
0.0132481391
0.0126728343
0.0120407078
0.0114302491
0.0114825013
0.010857286
0.0105780976
0.0106079296
0.0102235075
0.0097475339
0.0091469803
0.0091031579
0.0089834631
0.0088390912
0.0083835
0.00836578
0.0078131336
0.0080878237
0.0076635131
0.0078185222
0.0076041376
0.0072460159
0.0073998755
0.0076382502
0.0080252343
0.0073866491
0.0080853597
0.0076069599
0.0076429333
0.0068644188
0.0067108143
0.0060548676
0.0062005758
0.0053114711
0.0055107663
0.0057590726
0.0058247572
0.0063866339
0.0072045127
0.0075622624
0.0075940257
};
\end{axis}
\end{tikzpicture}
\caption{Average derivative of the LSTM output with respect to its input as a function of distance. The output is most sensitive to the closest words, but the tail of the distribution is fairly heavy, indicating that far-away words also have substantial impact.}
\label{fig:deriv}
\end{figure}
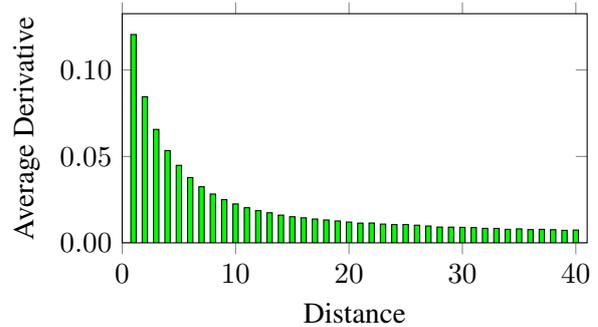

Figure~\ref{fig:deriv} illustrates how the average gradient norm is affected by the distance between the LSTM input and output.  As would be expected, the closest input vectors have the largest effect on the hidden state.  However, the tail of values is fairly heavy, with substantial gradient norms even for inputs 40 words away.  This shows that far-away inputs do have an effect on the LSTM representation.

\subsection{Truncation Analysis}
\label{sec:truncate}

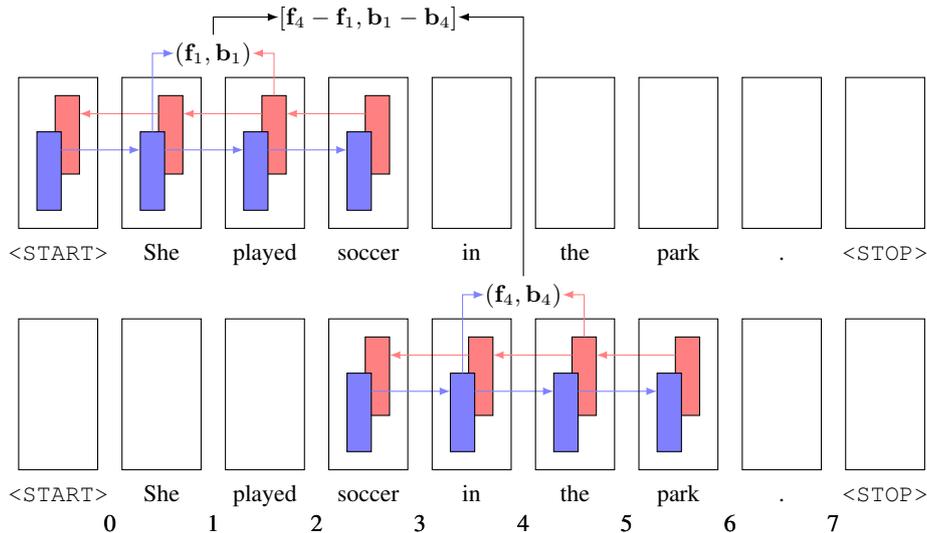
\begin{figure*}
\centering
\begin{tikzpicture}[every node/.style={font=\strut\small,inner sep=0},>=latex,scale=.8]
\newcommand\sentence{{\tt<START>},She,played,soccer,in,the,park,.,{\tt<STOP>}}
\newcommand\leftIndex{1}
\newcommand\rightIndex{4}
\newcommand\forwardColor{white!50!blue}
\newcommand\backwardColor{white!50!red}
\pgfmathsetmacro{\wordDistance}{1.7}
\pgfmathsetmacro{\indexOffsetV}{-.5}
\pgfmathsetmacro{\forwardOffsetV}{.7}
\pgfmathsetmacro{\lstmWidth}{.4}
\pgfmathsetmacro{\lstmHeight}{1.3}
\pgfmathsetmacro{\lstmSeparationH}{.3}
\pgfmathsetmacro{\lstmSeparationV}{.6}
\pgfmathsetmacro{\forwardArrowOffsetV}{\lstmHeight-\lstmSeparationV/2}
\pgfmathsetmacro{\backwardArrowOffsetV}{\lstmHeight-\lstmSeparationV/2}
\pgfmathsetmacro{\boxMargin}{.3}
\pgfmathsetmacro{\outputOffsetV}{.4}
\pgfmathsetmacro{\spanOffsetV}{.6}
\foreach \word [count=\finalIndex from -1] in \sentence;
\foreach \lstmPosition [count=\lstmIndex from 0] in {(0,0),(0,4)} {
  \foreach \word [count=\i from -1] in \sentence {
    \path \lstmPosition ++ ({\wordDistance*(\i)},0)
      node (word\i) {\word};
    \ifthenelse{\i>-1}{
      \node at ({\wordDistance*(\i-0.5)},\indexOffsetV) {\i};
    }{}
    \path
      \lstmPosition++(\wordDistance*\i-\lstmWidth/2+\lstmSeparationH/2,\forwardOffsetV+\lstmSeparationV)
      node (backwardLowerLeft) {}
      -- ++(\lstmWidth,\lstmHeight)
      node[midway] (backwardCenter) {}
      node (backwardUpperRight) {};
    \path
      \lstmPosition++(\wordDistance*\i-\lstmWidth/2-\lstmSeparationH/2,\forwardOffsetV)
      node (forwardLowerLeft) {}
      -- ++(\lstmWidth,\lstmHeight)
      node[midway] (forwardCenter) {}
      node (forwardUpperRight) {};
    \path
      (forwardCenter |- forwardUpperRight)
      node (forwardOut\i) {};
    \path
      (backwardCenter |- backwardUpperRight)
      node (backwardOut\i) {};
    \ifthenelse{\lstmIndex=1} {
      \pgfmathsetmacro{\startPosition}{\leftIndex-3}
      \pgfmathsetmacro{\endPosition}{\leftIndex+2}
    } {
      \pgfmathsetmacro{\startPosition}{\rightIndex-3}
      \pgfmathsetmacro{\endPosition}{\rightIndex+2}
    }
    \ifthenelse{\i>\startPosition \AND \i<\endPosition} {
      \draw[fill=\backwardColor] (backwardLowerLeft) rectangle (backwardUpperRight);
      \draw[fill=\forwardColor] (forwardLowerLeft) rectangle (forwardUpperRight);
    }{}
    \pgfmathsetmacro{\startPlusOne}{\startPosition+1}
    \ifthenelse{\i>\startPlusOne \AND \i<\endPosition}{
      \draw[<-,\forwardColor]
        (forwardLowerLeft)
        ++(0,\forwardArrowOffsetV)
        -- ++(-\wordDistance+\lstmWidth,0);
      \draw[->,\backwardColor]
        (backwardLowerLeft)
        ++(0,\backwardArrowOffsetV)
        -- ++(-\wordDistance+\lstmWidth,0);
    }{}
    \path
      (forwardLowerLeft) ++(-\boxMargin,-\boxMargin)
      node (boxLowerLeft) {};
    \path
      (backwardUpperRight) ++(\boxMargin,\boxMargin)
      node (boxUpperRight) {};
    \draw
      (boxLowerLeft) rectangle (boxUpperRight)
      node[midway] (boxCenter) {};
    \ifthenelse{\(\i=\leftIndex \AND \lstmIndex=1\) \OR \(\i=\rightIndex \AND \lstmIndex=0\)}{
      \path
        (boxCenter |- boxUpperRight) ++(-\wordDistance*0.5,\outputOffsetV)
        node (output\i) {$(\mathbf{f}_\i,\mathbf{b}_\i)$};
      \pgfmathtruncatemacro{\iPrev}{\i-1}
      \draw[->,\forwardColor] (forwardOut\iPrev.center) |- (output\i);
      \draw[->,\backwardColor] (backwardOut\i.center) |- (output\i);
    }{}
  }
}
\path
  (output\leftIndex) -- (output\rightIndex)
  node[midway] (outputMidpoint) {};
\path
  (outputMidpoint |- output\leftIndex) ++(0,\spanOffsetV)
  node (span) {$[\mathbf{f}_\rightIndex-\mathbf{f}_\leftIndex, \mathbf{b}_\leftIndex-\mathbf{b}_\rightIndex]$};
\draw[->] (output\leftIndex) |- (span);
\draw[->] (output\rightIndex) |- (span);
\end{tikzpicture}
\caption{An example of creating a truncated span representation for the span ``played soccer in'' with context size $k=2$.  This representation is used to investigate the importance of information far away from the fenceposts of a span.}
\label{fig:trunc-representation}
\end{figure*}

Next, we investigate whether information in the LSTM representation about far-away inputs is actually important for parsing performance.
To do so, we remove distant context information from our span encoding, representing spans by features obtained from LSTMs that are run on fixed-sized windows of size $k$ around each fencepost. Figure~\ref{fig:trunc-representation} illustrates this truncated representation.
Since the truncated representation also removes information about the size and position of the span in addition to the context words, we learn a position-dependent cell state initialization for each of the two LSTM directions to give a more fair comparison to the full LSTM.
The use of a fixed-sized context window is reminiscent of prior work by \citet{hall2014less} and \citet{durrett2015neural}, but here we use an LSTM instead of sparse features.
We train parsers with different values of $k$ and observe how their performance varies.
All other architecture details and hyperparameters are the same as for the original model.

The blue points in Figure~\ref{fig:truncation} show how the context size $k$ affects parser performance for $k\in\{2,3,5,10,20,30\}$. As with the derivative analysis, although most of the weight is carried by the nearby inputs, a nontrivial fraction of performance is due to context more than 10 words away.

\subsection{Word Order}

Now that we have established that long-distance information is important for parsing performance, we would like to know whether the order of the far-away words is important.
Is the LSTM capturing far-away structure, or is the information more like a bag-of-words representation summarizing the words that appear?

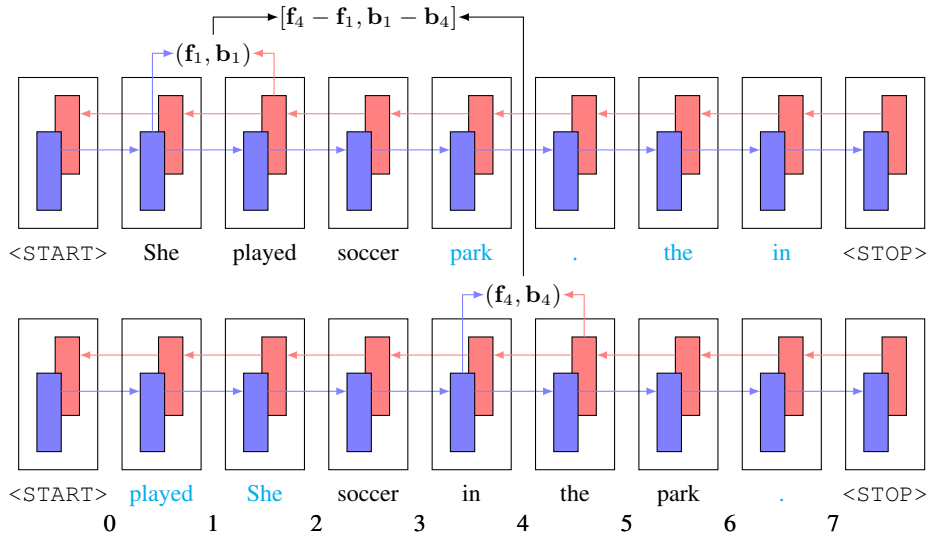
\begin{figure*}
\centering
\begin{tikzpicture}[every node/.style={font=\strut\small,inner sep=0},>=latex,scale=.8]
\newcommand\leftSentence{{\tt<START>},She,played,soccer,park,.,the,in,{\tt<STOP>}}
\newcommand\rightSentence{{\tt<START>},played,She,soccer,in,the,park,.,{\tt<STOP>}}
\newcommand\leftIndex{1}
\newcommand\rightIndex{4}
\newcommand\forwardColor{white!50!blue}
\newcommand\backwardColor{white!50!red}
\newcommand\shuffledWordColor{cyan}
\pgfmathsetmacro{\wordDistance}{1.7}
\pgfmathsetmacro{\indexOffsetV}{-.5}
\pgfmathsetmacro{\forwardOffsetV}{.7}
\pgfmathsetmacro{\lstmWidth}{.4}
\pgfmathsetmacro{\lstmHeight}{1.3}
\pgfmathsetmacro{\lstmSeparationH}{.3}
\pgfmathsetmacro{\lstmSeparationV}{.6}
\pgfmathsetmacro{\forwardArrowOffsetV}{\lstmHeight-\lstmSeparationV/2}
\pgfmathsetmacro{\backwardArrowOffsetV}{\lstmHeight-\lstmSeparationV/2}
\pgfmathsetmacro{\boxMargin}{.3}
\pgfmathsetmacro{\outputOffsetV}{.4}
\pgfmathsetmacro{\spanOffsetV}{.6}
\foreach \word [count=\finalIndex from -1] in \leftSentence;
\foreach \lstmPosition [count=\lstmIndex from 0] in {(0,0),(0,4)} {
  \ifthenelse{\lstmIndex=0}{\let\sentence\rightSentence}{\let\sentence\leftSentence}
  \foreach \word [count=\i from -1] in \sentence {
    \ifthenelse{\i>-1}{
      \node at ({\wordDistance*(\i-0.5)},\indexOffsetV) {\i};
    }{}
    \path
      \lstmPosition++(\wordDistance*\i-\lstmWidth/2+\lstmSeparationH/2,\forwardOffsetV+\lstmSeparationV)
      node (backwardLowerLeft) {}
      -- ++(\lstmWidth,\lstmHeight)
      node[midway] (backwardCenter) {}
      node (backwardUpperRight) {};
    \path
      \lstmPosition++(\wordDistance*\i-\lstmWidth/2-\lstmSeparationH/2,\forwardOffsetV)
      node (forwardLowerLeft) {}
      -- ++(\lstmWidth,\lstmHeight)
      node[midway] (forwardCenter) {}
      node (forwardUpperRight) {};
    \path
      (forwardCenter |- forwardUpperRight)
      node (forwardOut\i) {};
    \path
      (backwardCenter |- backwardUpperRight)
      node (backwardOut\i) {};
    \draw[fill=\backwardColor] (backwardLowerLeft) rectangle (backwardUpperRight);
    \draw[fill=\forwardColor] (forwardLowerLeft) rectangle (forwardUpperRight);
    \ifthenelse{\i>-1}{
      \draw[<-,\forwardColor]
        (forwardLowerLeft)
        ++(0,\forwardArrowOffsetV)
        -- ++(-\wordDistance+\lstmWidth,0);
      \draw[->,\backwardColor]
        (backwardLowerLeft)
        ++(0,\backwardArrowOffsetV)
        -- ++(-\wordDistance+\lstmWidth,0);
    }{}
    \ifthenelse{\lstmIndex=1} {
      \pgfmathsetmacro{\startPosition}{\leftIndex-3}
      \pgfmathsetmacro{\endPosition}{\leftIndex+2}
    } {
      \pgfmathsetmacro{\startPosition}{\rightIndex-3}
      \pgfmathsetmacro{\endPosition}{\rightIndex+2}
    }
    \ifthenelse{\(\i>\startPosition \AND \i<\endPosition\) \OR \i=-1 \OR \i=\finalIndex} {
      \path \lstmPosition ++ ({\wordDistance*(\i)},0)
        node (word\i) {\word};
    }{
      \path \lstmPosition ++ ({\wordDistance*(\i)},0)
        node[text=\shuffledWordColor] (word\i) {\word};
    }
    \path
      (forwardLowerLeft) ++(-\boxMargin,-\boxMargin)
      node (boxLowerLeft) {};
    \path
      (backwardUpperRight) ++(\boxMargin,\boxMargin)
      node (boxUpperRight) {};
    \draw
      (boxLowerLeft) rectangle (boxUpperRight)
      node[midway] (boxCenter) {};
    \ifthenelse{\(\i=\leftIndex \AND \lstmIndex=1\) \OR \(\i=\rightIndex \AND \lstmIndex=0\)}{
      \path
        (boxCenter |- boxUpperRight) ++(-\wordDistance*0.5,\outputOffsetV)
        node (output\i) {$(\mathbf{f}_\i,\mathbf{b}_\i)$};
      \pgfmathtruncatemacro{\iPrev}{\i-1}
      \draw[->,\forwardColor] (forwardOut\iPrev.center) |- (output\i);
      \draw[->,\backwardColor] (backwardOut\i.center) |- (output\i);
    }{}
  }
}
\path
  (output\leftIndex) -- (output\rightIndex)
  node[midway] (outputMidpoint) {};
\path
  (outputMidpoint |- output\leftIndex) ++(0,\spanOffsetV)
  node (span) {$[\mathbf{f}_\rightIndex-\mathbf{f}_\leftIndex, \mathbf{b}_\leftIndex-\mathbf{b}_\rightIndex]$};
\draw[->] (output\leftIndex) |- (span);
\draw[->] (output\rightIndex) |- (span);
\end{tikzpicture}
\caption{An example of creating a shuffled span representation for the span ``played soccer in'' with context size $k=2$.  The light blue words are outside the context window and are shuffled randomly.  Shuffled representations are used to explore whether the order of far-away words is important.}
\label{fig:shuffle-representation}
\end{figure*}

To test the importance of order, we train a parser where information about the order of far-away words is destroyed. As illustrated in Figure~\ref{fig:shuffle-representation}, we run a separate LSTM over the entire sentence for each fencepost, shuffling the input depending on the particular fencepost being represented. We randomly shuffle words outside a context window of size $k$ around the fencepost of interest, keeping words on the left and the right separate so that directional information is preserved but exact positions are lost.

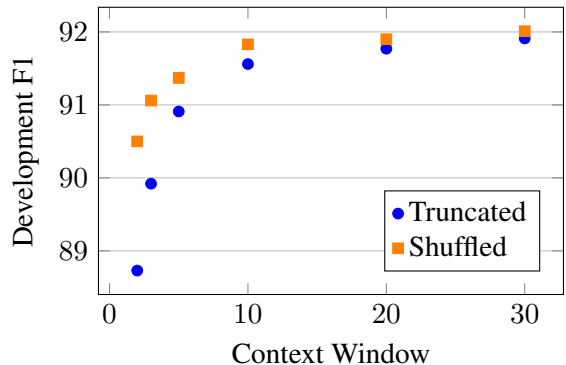
\begin{figure}[t]
\centering
\begin{tikzpicture}
\begin{axis}[
  width=\linewidth,
  height=.7\linewidth,
  xlabel=Context Window,
  ylabel=Development F1,
  ytick={87,...,92},
  ymajorgrids=true,
  legend style={at={(.95,.08)},anchor=south east},
  legend cell align={left},
]
\pgfplotstableread{
context truncated shuffled
2 88.73 90.5
3 89.92 91.06
5 90.91 91.37
10 91.56 91.83
20 91.77 91.9
30 91.91 92.01
}\dataTable
\addplot[only marks,blue] table[x=context,y=truncated] {\dataTable};
\addlegendentry{Truncated}
\addplot[only marks,orange,mark=square*] table[x=context,y=shuffled] {\dataTable};
\addlegendentry{Shuffled}
\end{axis}
\end{tikzpicture}
\caption{Development F1 as the amount of context given to the sentence-level LSTM varies.  The blue points represent parser performance when the LSTM is truncated to a window around the fenceposts, showing that far-away context is important. The orange points represent performance when the full context is available but words outside a window around the fenceposts are shuffled, showing that the order of far-away context is also important.}
\label{fig:truncation}
\end{figure}

The orange points in Figure~\ref{fig:truncation} show the performance of this experiment with different context sizes $k$. We observe that including shuffled distant words is substantially better than truncating them completely. On the other hand, shuffling does cause performance to degrade relative to the base parser even when the unshuffled window is moderately large, indicating that the LSTM is propagating information that depends on the order of words in far-away positions.

\subsection{LSTMs vs.\ Feedforward}
\label{sec:nolstm}

Finally, we investigate whether the LSTM architecture itself is important for reasons other than just the amount of context it can capture.
Like any architecture, the LSTM introduces particular inductive biases that affect what gets learned, and these could be important for parser performance.
We run a version of the truncation experiment from Section~\ref{sec:truncate} where we use a feedforward network in place of a sentence-level LSTM to process the surrounding context of each fencepost.
The input to the network is the concatenation of the word representations that would be used as inputs for the truncated LSTM, and the output is a vector of the same size as the LSTM-based representation.
As in Section~\ref{sec:truncate}, we wish to give our representation information about span size and position, so we also include a learned fencepost position embedding in the concatenated inputs to the network.
We focus on context window size $k=3$ for this experiment.
We search among networks with one, two, or three hidden layers that are one, two, or four times the size of the LSTM hidden state.

Of all the feedforward networks tried, the maximum development performance was 83.39 F1, compared to 89.92 F1 for the LSTM-based truncation.  This suggests that some property of the LSTM makes it better suited for the task of summarizing context than a flat feedforward network.

\section{Related Analysis Work}

Here we review other works that have performed similar analyses to ours in parsing and other areas of NLP.  See Section~\ref{sec:model} for a description of how our parser is related to other parsers.

Similar to our independent span prediction in Section~\ref{sec:independent}, several works have found that their models still produce valid outputs for the majority of inputs even after relaxing well-formedness constraints.
In dependency parsing, \citet{zhang2017dependency} and \citet{chorowski2016read} found that selecting dependency heads independently often resulted in valid trees for their parsers (95\% and 99.5\% of outputs form trees, respectively).
In constituency parsing, the parser of \citet{vinyals2015grammar}, which produced linearized parses token by token, was able to output valid constituency trees for the majority of sentences (98.5\%) even though it was not constrained to do so.

Several other works have investigated what information is being captured within LSTM representations.  \citet{chawla2017investigating} performed analysis of bidirectional LSTM representations in the context of named entity recognition.  Although they were primarily interested in finding specific word types that were important for making decisions, they also analyzed how distance affected a word's impact.
\citet{xing2016stringbased} and \citet{linzen2016assessing} perform analysis of LSTM representations in machine translation and language modeling respectively to determine whether syntactic information is present.  Some of their techniques involve classification of features from LSTM hidden states, similar to our analysis in Sections~\ref{sec:parentclass} and~\ref{sec:wordfeat}.

In Section~\ref{sec:nolstm}, we found that replacing an LSTM with a feedforward network hurt performance.
Previously, \citet{chelba2017ngram} had similar findings in language modeling, where using LSTMs truncated to a particular distance improved performance over feedforward networks that were given the same context.

\section{Conclusion}

In this paper, we investigated the extent to which information provided directly by model structure in classical constituency parsers is still being captured by neural methods.
Because neural models function in a substantially different way than classical systems, it could be that they rely on different information when making their decisions.
Our findings suggest that, to the contrary, the neural systems are learning to capture many of the same knowledge sources that were previously provided, including the parent-child relations encoded in grammars and the word features induced by lexicons.

\section*{Acknowledgments}

This work is supported by the DARPA Explainable Artificial Intelligence (XAI) program and the UC Berkeley Savio computational cluster. The second author is supported by an NSF Graduate Research Fellowship.


\bibliography{naaclhlt2018}
\bibliographystyle{acl_natbib}

\clearpage

\appendix
\onecolumn

\section{Model Hyperparameters and Training Details}
\label{appendix:hyperparameters}

\begin{table}[h]
\centering
\begin{tabular}{c|cc}
Component & Dimensions & Layers \\ \hline
Word Embeddings & 100 \\
Character Embeddings & 50 \\
Character LSTM & 100 & 1 \\
Sentence LSTM & 250 & 2 \\
Label Feedforward Network & 250 & 1 \\
\end{tabular}
\caption{The sizes of the components used in our model.}
\label{table:parameters}
\end{table}

Our model hyperparameters are summarized in Table~\ref{table:parameters}. We train using the Adam optimizer \citep{kingma2014adam} with its default hyperparameters for 40 epochs. We evaluate on the development set 4 times per epoch, selecting the model with the highest overall development performance as our final model. When performing a word embedding lookup during training, we randomly replace words by the {\tt <UNK>} token with probability $1/(1+\text{freq}(w))$, where $\text{freq}(w)$ is the frequency of a word $w$ in the training set. We apply dropout with probability 0.4 before and inside each layer of each LSTM. Our system is implemented in Python using DyNet \citep{neubig2017dynet}.

\section{Character LSTM Word Feature Classification}
\label{appendix:charlstm}

\begin{table}[h]
\centering
\hfill
\begin{tabular}[t]{ccc}
& Majority & Char-LSTM \\
Binary Feature & Class & Classifier \\ \hline
all-letters & 77.22\% & 99.77\% \\
has-letter & 89.18\% & 99.97\% \\
all-lowercase & 56.95\% & 99.95\% \\
has-lowercase & 85.85\% & 99.90\% \\
all-uppercase & 96.68\% & 99.90\% \\
has-uppercase & 67.77\% & 99.97\% \\
all-digits & 98.38\% & 99.99\% \\
has-digit & 87.90\% & 99.91\% \\
all-punctuation & 99.93\% & 99.98\% \\
has-punctuation & 79.04\% & 99.75\% \\
has-dash & 88.89\% & 99.95\% \\
has-period & 92.55\% & 99.95\% \\
has-comma & 98.02\% & 99.97\% \\
\end{tabular}
\hfill
\begin{tabular}[t]{ccc}
& Majority & Char-LSTM \\
Binary Feature & Class & Classifier \\ \hline
suffix = ``s'' & 82.65\% & 99.99\% \\
suffix = ``ed'' & 92.52\% & 99.98\% \\
suffix = ``ing'' & 93.26\% & 99.95\% \\
suffix = ``ion'' & 97.75\% & 99.93\% \\
suffix = ``er'' & 96.42\% & 99.97\% \\
suffix = ``est'' & 99.63\% & 99.98\% \\
suffix = ``ly'' & 97.56\% & 99.99\% \\
suffix = ``ity'' & 99.30\% & 99.94\% \\
suffix = ``y'' & 92.97\% & 99.93\% \\
suffix = ``al'' & 98.48\% & 99.92\% \\
suffix = ``ble'' & 99.30\% & 99.90\% \\
suffix = ``e'' & 89.57\% & 99.99\% \\
\end{tabular}
\hfill\hfill 
\caption{Classification accuracy for various binary word features using the character LSTM representations for words induced by a pre-trained parser. Performance substantially exceeds that of a majority class classifier in all cases, reaching 99.7\% or higher for all features. The majority class is {\tt True} for the first four features in the left column and {\tt False} for the rest.}
\label{table:word-features}
\end{table}

\end{document}